\documentclass{article}

\pdfoutput=1
\usepackage{arabtex}
\usepackage{utf8}
\setcode{utf8}

\usepackage{PRIMEarxiv}
\usepackage[utf8]{inputenc} 
\usepackage[T1]{fontenc}    
\usepackage{hyperref}       
\usepackage{url}            
\usepackage{booktabs}       
\usepackage{amsfonts}       
\usepackage{nicefrac}       
\usepackage{microtype}      
\usepackage{float}          
\usepackage{fancyhdr}       
\usepackage{graphicx}       
\graphicspath{{media/}}     

\title{ArmanEmo: A Persian
Dataset for Text-based Emotion Detection}

\author{
  Hossein Mirzaee \\
  Department of Chemical Engineering \\
  Amirkabir University of Technology \\
  Tehran, Iran\\
  \texttt{hmirzaee@aut.ac.ir} \\
  \And
  Javad Peymanfard \\
  School of Computer Engineering \\
  Iran University of Science and Technology \\
  Tehran, Iran \\
  \texttt{javad\_peymanfard@comp.iust.ac.ir} \\
  \And
  Hamid Habibzadeh Moshtaghin \\
  Faculty of Management and Accounting \\
  Allameh Tabataba'i University \\
  Tehran, Iran\\
  \texttt{h.habibzadeh@atu.ac.ir} \\
   \And
   Hossein Zeinali \\
   Department of Computer Engineering \\
   Amirkabir University of Technology \\
   Tehran, Iran\\
   \texttt{hzeinali@aut.ac.ir} \\
}

\begin{document}
\maketitle

\begin{abstract}
With the recent proliferation of open textual data on social media platforms, Emotion Detection (ED) from Text has received more attention over the past years. It has many applications, especially for businesses and online service providers, where emotion detection techniques can help them make informed commercial decisions by analyzing customers/users' feelings towards their products and services. In this study, we introduce ArmanEmo, a human-labeled emotion dataset of more than 7000 Persian sentences labeled for seven categories. The dataset has been collected from different resources, including Twitter, Instagram, and Digikala \footnote{an Iranian e-commerce company.} comments. Labels are based on Ekman's six basic emotions (Anger, Fear, Happiness, Hatred, Sadness, Wonder) and another category (Other) to consider any other emotion not included in Ekman's model. Along with the dataset, we have provided several baseline models for emotion classification focusing on the state-of-the-art transformer-based language models. Our best model achieves a macro-averaged F1 score of 75.39 percent across our test dataset. Moreover, we also conduct transfer learning experiments to compare our proposed dataset's generalization against other Persian emotion datasets. Results of these experiments suggest that our dataset has superior generalizability among the existing Persian emotion datasets. ArmanEmo is publicly available for non-commercial use at \url{https://github.com/Arman-Rayan-Sharif/arman-text-emotion}.
\end{abstract}


\section{Introduction}
Emotion expression and detection play important roles in human social and professional life. They are closely related to our cognitive abilities and communication skills, profoundly shaping the range and quality of our experiences and social interactions \cite{dolan2002emotion}. Consequently, emotional intelligence is one of the essential abilities to move from narrow to general human-like Artificial Intelligence \cite{ragheb2019attention}, \cite{poria2019emotion}. Emotion Detection (ED) is an active area of research to enable machines to understand human emotions effectively.

Emotion detection and analysis have been widely researched and have many applications in several fields such as neuroscience, psychology and behavioral sciences, computer science, and computational linguistics \cite{canales2014emotion}, \cite{strapparava2008learning}. In business and online commercial activities, companies and organizations are eager to analyze their customer’s emotions toward the company and their products so they can adapt their behavior, services, and products to the customer expectations and hence, ensure the business growth \cite{acheampong2020text}. Due to the current expansion of the social platforms and the recent proliferation of open conversational data, Emotion Recognition in Conversation (ERC) has become another attractive subfield of ED, which in turn has many potential applications in important downstream tasks such as emotionally intelligent chatbots, emotion-aware health-care, and educational systems, opinion mining and recommender systems \cite{poria2019emotion, zhou2018emotional, huang2019ana, majumder2019dialoguernn, lin2019moel, rashkin2018towards}.

Emotions can be expressed verbally (through words and tone of voice) or nonverbally (facial expressions or body language), and they, therefore, can be recognized in many sources like images and videos \cite{boubenna2018image}, sounds \cite{chatterjee2019human}, texts \cite{calefato2017emotxt}. With respect to ED from voice/speech, images, videos, and other multimodal data, ED from text is more challenging and has been studied less than the other approaches \cite{acheampong2020text}. This is because the text alone may be insufficient to recognize emotions \cite{huang2018automatic}. Even for humans, knowing the context of a conversation is essential to recognize the emotions from a text, given the lack of other data such as tone of voice and facial expressions in this case \cite{huang2019ana}. Diversity in the style of writing, formal and informal styles \cite{sadeghi2021automatic}, the usage of figurative language, like sarcasm \cite{chatterjee2019understanding}, detecting emotions from short texts, and presence of grammatical errors \cite{acheampong2020text} are some of the existing difficulties, which make the ED from text more challenging than it might seem at first glance. Regardless, the ever-increasing rise in the amount of available textual data in social media and its vast potential applications have made the ED from text an important domain to study \cite{huang2019ana}.

The goals of this paper are to:

\begin{itemize}
	\item Conduct a review study on Text-based Emotion Detection (especially  Emotion Detection from Persian Text)
	\item Contribute to Persian Text ED:
	\begin{itemize}
	\item By introducing a new dataset for Emotion Detection 
	\item By introducing strong baseline language models
	\end{itemize}
\end{itemize}

The rest of this paper is organized as follows: in \autoref{sec:2} the models of emotion and different approaches to emotion detection from text have been discussed, and a few studies regarding emotion detection from Persian text has been reviewed. \autoref{sec:3} provides our main contributions to emotion detection from Persian Text ED by introducing ArmanEmo, our new dataset, and the procedures we have followed to generate it. This section also presents several baseline models trained on ArmanEmo and their performance, along with various experiments to demonstrate the high quality of the dataset. Finally, \autoref{sec:4} concludes the paper.

\section{Text Based Emotion Detection}
\label{sec:2}

In this section, we first discuss emotion models used in previous studies related to Emotion Detection. We then review different approaches used to detect emotion from text. Finally, in the last part, we provide a brief review of previous works in emotion detection from Persian text.

\subsection{Emotion Models}

Fundamental to the development of Emotion Detection systems are emotion models that determine how various human affects are represented. Affect studies have been done in a variety of fields such as neuroscience, psychology, and cognitive sciences. According to research in psychology, there are three distinguished approaches for emotion modeling which are Categorical, Dimensional, and Appraisal-based approaches \cite{grandjean2008conscious}, \cite{gunes2010automatic}. However, the first two are the most important and often used approaches in emotion detection studies \cite{acheampong2020text, canales2014emotion}.

The categorical approaches suggest that human beings have a limited number of psychological and biologically basic emotions that are universally recognized \cite{ekman2003unmasking}. So, they involve classifying emotions into distinct classes or categories. The most commonly adopted approach within the categorical or discrete approaches is that of Paul Ekman \cite{dalgleish2000handbook}, which proposes that the six basic human emotions are ANGER, DISGUST, FEAR, HAPPINESS, SADNESS, and WONDER. Plutchik adds two more emotions to Ekman’s set of emotions, meaning TRUST and ANTICIPATION, so the whole set can be organized into four bipolar subsets: joy vs. sadness, anger vs. fear, trust vs. disgust, and surprise vs. anticipation \cite{plutchik1984emotions}.

While the majority of the automatic Emotion Systems have been based on categorical approaches, some researchers argue that any small set of basic emotion classes may not reflect some of the non-basic and complex affective states of human communications like thinking or depression \cite{gunes2011emotion}. Hence, they suggest representing emotions in a dimensional form. The main idea is that since the emotions are not independent, we need to systematically capture their relations with each other by placing them in a 1D, 2D, or 3D spatial space \cite{gunes2010automatic}. The most widely used dimensional model is Russell’s Circumplex Model of Affect \cite{russell1980circumplex}, suggesting that emotions are distributed in a two-dimensional circular space:

\begin{itemize}
\item Valence dimension: which differentiates emotions by Pleasantness and Unpleasantness
\item Arousal dimension: which indicates how excited or apathetic an emotion is
\end{itemize}

Most computational approaches for Emotion Detection are based on categorical emotion models, mainly because of their simplicity and familiarity. Nonetheless, since they are limited to a fixed number of emotions, they may not cover non-basic, mixed and complex emotional states \cite{yu2004detecting}. On the other hand, Dimensional approaches are more helpful in depicting these subtle emotions that differ only slightly. They are also highly recommended when the goal is to measure similarities between emotions \cite{acheampong2020text}. However, fitting all basic emotions in the dimensional space is not feasible since some become indistinguishable, and some may lie outside the space \cite{gunes2010automatic}. As we can see, both approaches have advantages and disadvantages, and none of them is superior to another in all situations. The selection of an emotion model depends on the end goal of developing an Emotion Detection system and the set of emotions we expect from the system to detect \cite{canales2014emotion}.

\subsection{Emotion Detection Approaches}

Among different computational approaches for Emotion Detection from text, three of them are currently more dominant: rule-based, Machine Learning-based, and hybrid approaches \cite{binali2010computational}. The rule-based approaches are based on following grammatical and logical rules to classify the text into different emotion categories. Such rules can be determined using linguistic, statistical, or computational concepts. One of the most widely used rule-based approaches is keyword recognition which relies on finding occurrences of predefined keywords in a given text at the sentence level. This keyword list or dictionary is prepared with the semantic labels of emotion such as sadness or anger, and sometimes it also indicates the intensity of the emotion. Once the keyword is identified within the sentence, emotion intensity measurement and negation checking are performed, and finally, an emotional label is assigned to the sentence \cite{sadeghi2021automatic}. The keyword-based approaches are straightforward and intuitive, yet, they face some challenges. Limited and domain-specific keyword lists and poor pre-processing heavily affect the emotion detection performance \cite{acheampong2020text, hancock2007expressing}.

In ML-based approaches, trained classifiers are used to automatically assign an emotional label to the input sentence. Supervised learning approach is one type of ML-based technique that relies on extensive training data annotated with emotional tags or labels. Trained on the training set to learn how to classify a given text into an emotion label, the supervised classification algorithm infers a function, which eventually can be used to map any new unseen examples into emotion labels. Unlike keyword-based approaches, supervised learning approaches are more adaptable to changes in the domain since they can effectively and quickly learn new features from the text \cite{binali2010computational}. However, to train such classifiers, a large labeled dataset is needed, which may be a tedious and time-consuming task.

Among classical supervised learning algorithms, SVM has been widely used for Emotion Detection from text \cite{alswaidan2020survey}. Given the meaningful feature set (obtained after performing some preprocessing, linguistic and statistical analysis on the input texts, and emotion labels), the SVM algorithm outputs an optimal hyperplane in multidimensional space to separate different emotion labels. Deep Learning algorithms are the other supervised ML-based approaches that have recently received considerable attention in Emotion Detection from text. It is argued that the deep layers of these algorithms enable them to capture the variations in the meaning of a word depending on its context \cite{huang2019emotionx}. Different variations of Recurrent Neural Networks, such as LSTM and GRU, Convolutional Neural Networks, and Transformer-based models are some techniques that have been used in Emotion Detection from text \cite{sadeghi2021automatic}. Implementation of some of these architectures has led to state-of-the-art results in Emotion Detection \cite{acheampong2020text}.

Unsupervised learning algorithms are the other type of ML-based approaches. In unsupervised learning approaches, instead of training a model on labeled or annotated data, the goal is to find some hidden structures in unlabeled data upon which models for emotion detection can be built \cite{canales2014emotion}.

In hybrid approaches, rule-based approaches and ML-based approaches are consolidated into a new solution that has the strengths of both approaches while alleviating their associated weaknesses. The idea is that one can yield more accurate results in emotion detection by implementing an ensemble of classifiers and adding knowledge-rich linguistic information from dictionaries \cite{binali2010computational}. It should be noted, however, that the performance enhancement a hybrid method can provide heavily relies on the particular types of classification methods used \cite{acheampong2020text}.

\subsection{Previous works in Emotion Detection from Persian Text}

Very few studies have been conducted on emotion detection from Persian text. In one of these studies \cite{sadeghi2021automatic}, a hybrid approach, based on the combination of cognitive features and Word2Vec embeddings, is proposed to achieve better performance in emotion detection. The emotional constructions, keywords, and parts of speech are utilized to construct the cognitive features. The resulting embedding vector is then used in a GRU network. It is suggested that this hybrid method outperforms the methods that rely solely on linguistic information or deep learning approaches. However, in general, the linguistic rules cannot be used in many practical applications. For instance, it is challenging to extract cognitive features in social media where the usage of colloquial and informal language is dominant. It is noteworthy that, to perform the experiments, the authors have sampled 23,000 sentences from Bijan Khan's corpus, making sure that each sentence carries only one emotion among five basic emotions (fear, happiness, sadness, anger, and surprise). Unfortunately, this dataset is not publicly available.

In another study, EmoPars, a dataset containing 30,000 Tweets, has been published \cite{sabri2021emopars}. EmoPars is the largest source of Persian text-based emotions; however, it has problems that make this large amount of labeled data inefficient. It is labeled through crowd-sourcing, in which each sentence has been given to 5 voters, and the final labels are to be acquired using maximum voting. Moreover, a significant number of selected tweets do not contain any emotions and are neutral. Only 23\% of the total dataset, including 7026 samples, are labeled with emotions. To evaluate the quality and generalizability of this dataset against ArmanEmo, we perform an experiment, comparing the result of the best model in our baselines when it is trained on Emopars against when it is trained on our dataset. This experiment is described in detail in section \ref{subsubsec:333}.

\section{Our Study}
\label{sec:3}

This section presents our contributions toward emotion detection from Persian text. First, we discuss the procedures we have followed to collect and annotate our Emotion dataset. Statistics regarding ArmanEmo are also provided. In the second part of this section, we introduce our baseline models and describe the preprocessing steps we have followed. Finally, in the third part, we present and discuss the results of all the experiments.

\subsection{Our New Emotion Dataset}
\label{subsec:31}

In this section, we demonstrate our new emotion dataset along with its statistics and describe the methodology we have followed for data collection and annotation. 

\subsubsection{Data Collection}

We have used different resources to collect the raw textual data we needed for this study. Of the most important resources that are recently receiving more attention in the field of emotion classification studies is the large text corpus coming from social media platforms. This is because individuals are increasingly using these online platforms to communicate their ideas and feelings about various topics. Since one of the main goals of this project is to evaluate the emotions towards different social and political topics, we have included Persian tweets published on Twitter, one of the widely used social media platforms in Iran. However, to make ArmanEmo more general and representative, we have also used two other resources. Along with a text corpus from users’ comments on Instagram, we have included customers’ comments on Digikala (an online shopping platform) in the dataset. Table \ref{tab:resources} summarises the resources we have used and the related details about them.

\begin{table}[H]
 \caption{Resources used to collect textual data}
  \centering
  \begin{tabular}{llll}
    \toprule
       & Persian Tweets  & Instagram Comments  & Digikala Comments \\
    \midrule
    Collection Method & Tweeter's official API & Facebook Graph API & Polite Crawling of the Website \\
    \midrule
    Collection Period & 2017 - mid 2018 & mid 2017 - mid 2018 & mid 2018 \\
    \midrule
    \# Raw Data & 1.5 M & 1 M & 50 K \\
    \midrule
    \# Data Used for &   &   &  \\
    Manual Annotation & 3.5 K & 3 K & 1 K \\
    \midrule
    \# Data Used for &  &  &  \\
    Automatic Annotation & 4.5 K & - & - \\
    \bottomrule
  \end{tabular}
  \label{tab:resources}
\end{table}

\subsubsection{Data Annotation}

After data collection, we took the following steps to improve the data quality before the annotation process:

\begin{enumerate}

  \item For the data coming from each of the above-mentioned resources, after calculating the distribution of sentences according to their lengths (in characters), we removed all the sentences whose lengths felt outside a specific range. We also eliminated sentences containing specific user IDs or links.
  
  \item In order to loosely control the class balance in our target dataset, we used a heuristic technique to weakly label each given sentence in the dataset before presenting them to our annotators for labeling. These "weak labels" are not final and are not guaranteed to be consistent with the labels selected by our annotators (weak labels were hidden from our annotators to avoid any bias in the labeling system). However, they are still helpful tools during the labeling process to pre-select sentences from rare classes more often so that the resulting dataset gets as much balanced as possible. The heuristic technique we used to generate weak labels relies on the NRC Word-Emotion Association Lexicon \cite{Mohammad13}, a dataset containing a list of emotional words and their association with basic emotions. We pre-classified the sentences coming from each of the three resources according to the association between their words and each of the emotion classes. We also used a weighted random selection method based on the Term-Frequency of emotional words in the sentences to make sure that those sentences containing emotional words with higher Term-Frequency are more likely to be selected and shown to the annotators during the labeling process. 
  
  \item In the end, we chose 12000 sentences to be labeled in the next step. Before handing these to our annotators (in the manual annotation process), we removed any emojis in these sentences to make sure they categorize the emotions in each given sentence only based on its text.
  
\end{enumerate}

Here we describe the annotation procedure, which is a mix of manual and automatic steps. Manual data annotation was performed on 7500 sentences (out of 12000 selected sentences) through the application of a Telegram Bot developed for this project using tools like official Telegram APIs and MySQL. At the beginning of their interaction with this bot, users were first introduced to our specially designed instructions for data annotation. Then, their performances in following our rules and instructions were evaluated and measured against our existing labeled test set. The test set containing 250 sentences had already been manually labeled and validated by ten annotators. Among the 35 users who participated in the data labeling, we selected 12 of them with the highest scores as our final annotators. The data annotation was done in March 2019.

Based on our instructions, given a sentence, annotators have a few options to select. If annotators are sure that the sentence has only one specific emotion among our six predefined categories, they would need to choose that emotion as the label of the sentence. If they decide that the sentence carries no specific emotion or emotions other than our predefined classes, they will select "Other." Otherwise, if they cannot confidently assign a specific feeling to the given sentence, they should select the "Unknown" option.

By the end of the labeling loop, each given sentence needed to be classified into the same emotion category by three different annotators considering that it had not been shown to more than five annotators. In other words, if the sentence had already been shown to five annotators without being finalized, it would be removed from the labeling procedure and will not be shown to other annotators until the last step of the labeling (to be explained later). To speed up the labeling process, we did not hand the sentences to the annotators all at once. Instead, we prioritize the introduction of the sentences to the annotators based on their previous labels so that those sentences that needed less labeling effort to be finalized (and temporarily removed from the labeling loop) get displayed to the annotators sooner than the others. For example, a sentence labeled by three annotators without being finalized is of more priority with respect to the sentence that has been classified once (by only one annotator).

In order to control the quality of labels during the labeling process, we also used a data dashboard to capture and display the instant information related to our annotation system. The aim for the development and application of such a system was:

\begin{enumerate}
   \item \textbf{To balance the class distribution in the final labeled dataset}:
   Observing the class distribution of the labeled data during the annotation process, we tried to update the selection weights (by which new weakly-labeled sentences were introduced to our annotators) in reverse proportion to the frequency of each label in our labeled data. We wanted to present new sentences with rare labels (i.e., their corresponding weak-label) to our annotators more often so that we have a final dataset that is as much balanced as possible. It is worth noting that the weak labels are still weak, and they were likely to be rejected by our annotators. Due to this reason, our final labeled dataset is not perfectly balanced.
   
   \item \textbf{To analyze the performance of the annotators}:
   During the annotation process, we evaluated the performance of each annotator based on their contribution to labeling the finalized examples. The more they had involved in finalizing the labels for input data, their performance score was higher. Among the twelve annotators, three of them got the highest score whose judgment and skills were used in labeling the remaining examples that had not been finalized in previous steps. One of the annotators, on the other hand, had a very low and unacceptable score and was eliminated from the annotation step.

\end{enumerate}

By the end of the manual annotation steps, we had  4700 sentences labeled by our annotators. We randomly split this data into primary training (>3500 samples) and testing sets (>1100 samples). We then fine-tuned ParsBert Language Model \cite{farahani2021parsbert} on this training set (with the same hyperparameters described in section \ref{subsec:32}). We used this classifier to label the second parts of our selected input sentences (4500 samples from Persian Tweet, \ref{tab:resources}). Among these sentences, 2000 samples with lower confidence were filtered out from our annotation process. An annotator then manually checked the labels of the remaining 2500 sentences and changed them whenever needed. These 2500 samples were then added to our training set.

\subsubsection{Dataset Statistics}

\begin{figure}[H]
\centering
\includegraphics[width=8cm]{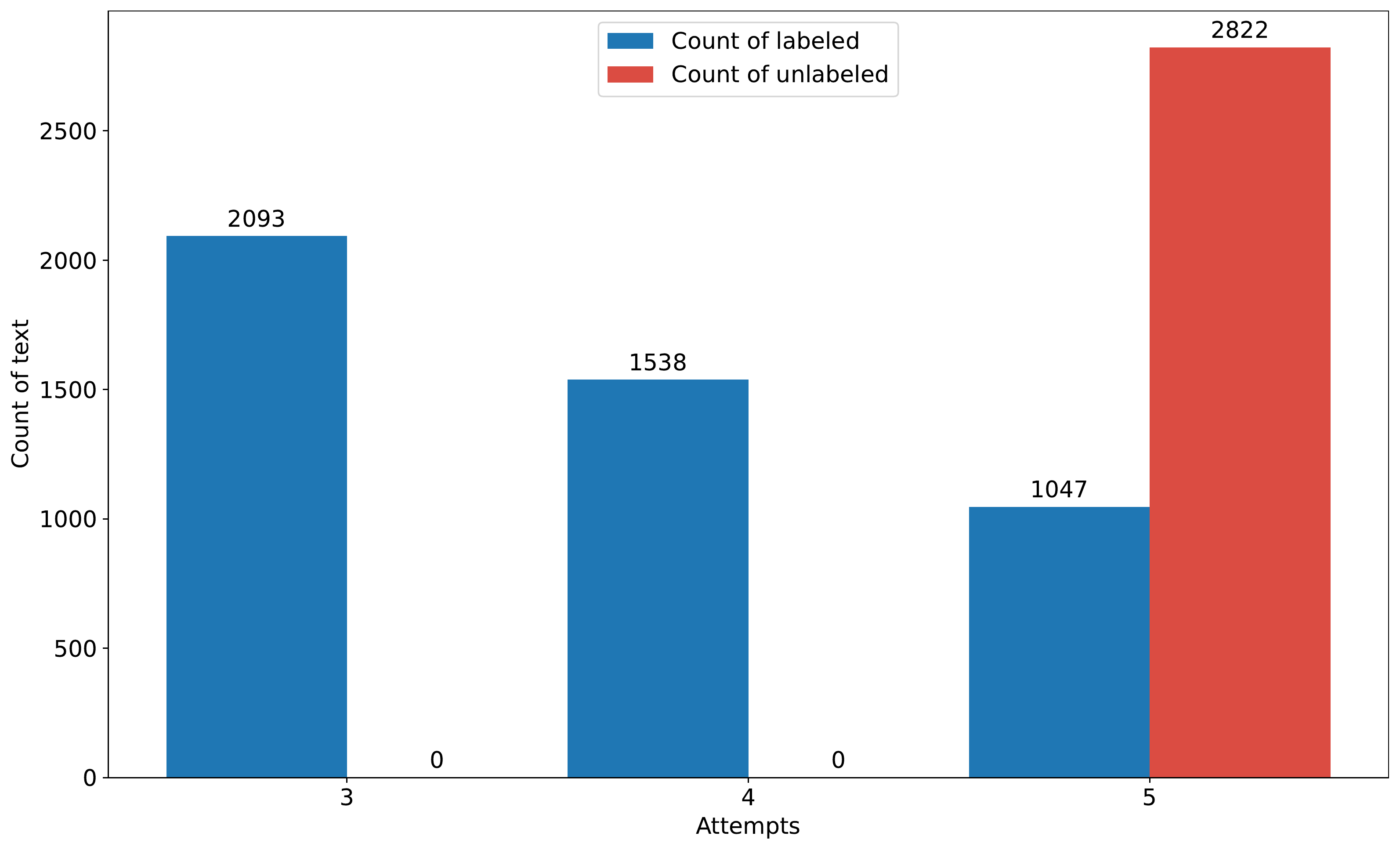}
\caption{Count of labeled and unlabeled data per attempts}
\label{fig:labelsperattempts}
\end{figure}

As mentioned before, we selected 7500 sentences to be manually labeled as one of the emotion classes or two other categories, i.e., "Unknown" and "Other." Figure \ref{fig:labelsperattempts} shows the count of labeled and unlabeled data per attempt (each iteration of the annotation process). We observed that the labels of 38 percent of these 7500 sentences ended up undecided because they did not label as one of the emotion classes after reviewing by the fifth annotator. This observation is another piece of evidence showing that emotion classification from text is a challenging task even for human annotators.

\begin{figure}[H]
\centering
\includegraphics[width=8cm]{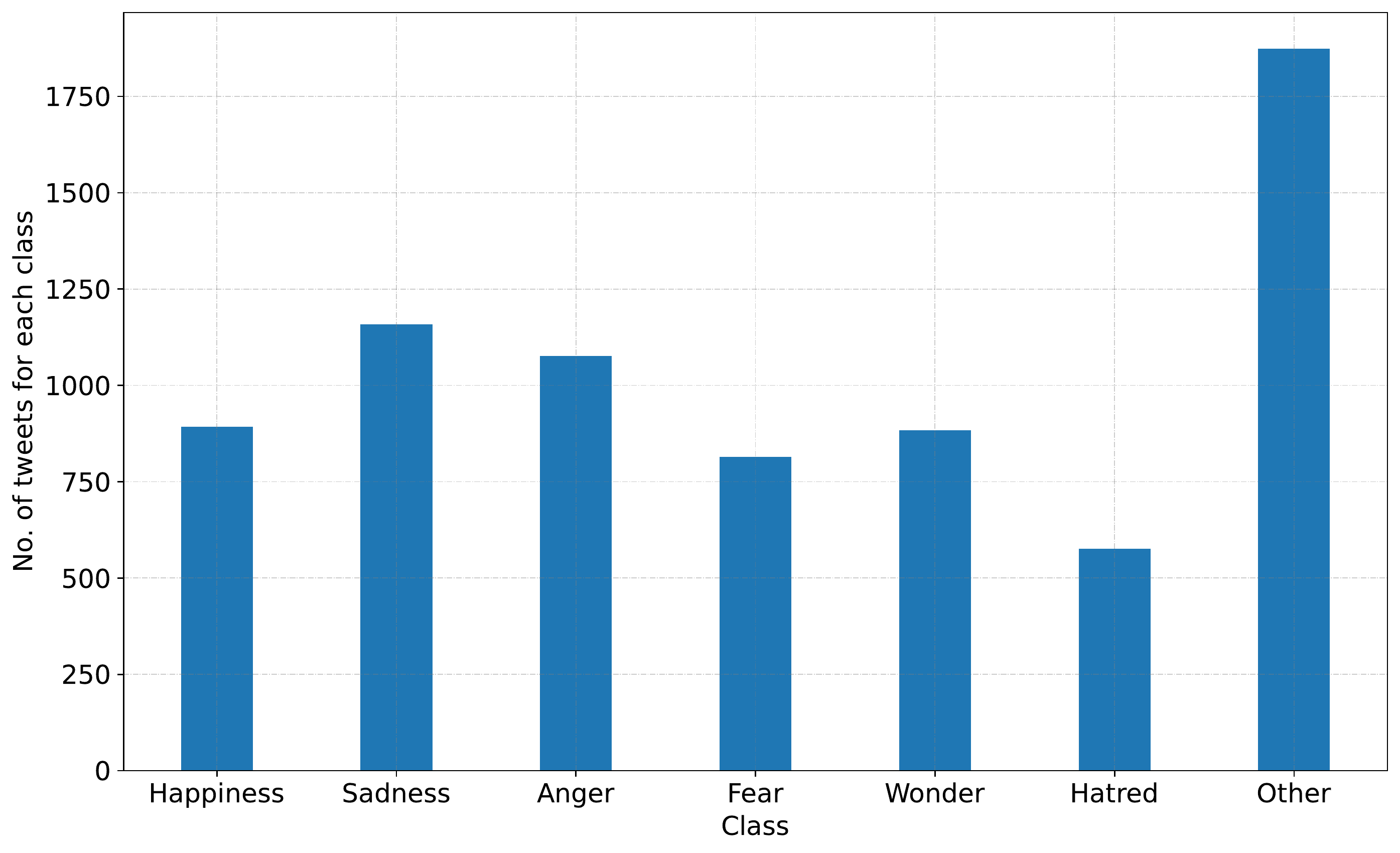}
\caption{No. of Tweets for Each Class}
\label{fig:tweetsperclass}
\end{figure}

In figure \ref{fig:tweetsperclass} the label frequency for all the (manually and automatically) labeled data is provided. Twenty-five percent of the final labeled sentences had been classified as "Other." Moreover, some data was labeled as "Unknown" by the end of the annotation process, which we decided to remove from the dataset.

\subsection{Modeling}
\label{subsec:32}

In this section, we provide baseline models for emotion classification over ArmanEmo, which use transfer learning. The fact is that supervised learning is difficult to apply to NLP problems, including emotion detection, since labels are costly. Here is where transfer learning comes into play. Transfer learning from pre-trained deep neural Language Models (LMs) towards downstream language problems has led to a state-of-the-art performance in several NLP tasks in recent years \cite{kant2018practical, ying2019improving, durrani2021transfer}. Deep LMs can be efficiently pre-trained in an unsupervised way over very large unlabeled datasets. In this way, the resulting LMs will capture rich and non-trivial linguistic knowledge, making them suitable to be transferred to a subsequent target domain and task through supervision. To be adaptive to a target domain, the pre-trained LMs need to be fine-tuned by a small amount of labeled data from that domain.

As one of our baseline models, we take advantage of a pre-trained language model for Persian, known as ParsBERT. ParsBERT is a monolingual language model based on Bidirectional Encoder Representation Transformer (BERT) architecture. Farahani et al. \cite{farahani2021parsbert} have shown that the ParsBERT model outperforms the multilingual BERT and previous models in several Persian NLP downstream tasks, including text classification and sentiment analysis. Lighter than the multilingual BERT, ParsBERT has been trained on a larger and more diverse (in terms of the range of topics and style of writing) set of pre-trained Persian datasets.

We also use two variations of a model known as XLM-RoBERTa as our other baseline models. XLM-RoBERTa is another transformer-based masked language model pre-trained on text in 100 languages. This multilingual language model has led to state-of-the-art performance on cross-lingual classification, sequence labeling, and question answering, outperforming multilingual BERT (mBERT) on various cross-lingual benchmarks \cite{liu2020multilingual}. Although it is known that ParsBERT as a monolingual language model outperforms multilingual BERT on various tasks in Persian language, we decided to compare the performance of XLM-RoBERTa variations (namely XLM-RoBERTa-base and XLM-RoBERTa-large) against ParsBERT on emotion detection from Persian text.

Another model included in our baselines is XLM-EMO \cite{bianchi2022xlm}, a multilingual emotion detection model for social media text. It is essentially XLM-T fine-tuned on datasets for emotion detection in 19 different languages. XLM-T itself is a fine-tuned version of XLM-RoBERTa-base on Twitter data \cite{barbieri2021xlmtwitter}. In developing XLM-EMO model, emotion labels of each dataset are mapped to a common set, namely joy, anger, fear, and sadness. It has been shown that this multilingual emotion model is competitive against language-specific baselines in zero-shot settings. It is specifically developed to help low-resource languages that still do not have a dataset for emotion detection. In our study, we evaluate the performance of this model in three different settings. In the zero-shot setting, we want to assess this model's accuracy in emotion detection from Persian text without being trained on our emotion dataset. To have a basis of comparison for the zero-shot learning in the first setting, we also fine-tune XLM-EMO on our train set and evaluate it on our test set. To perform these two studies, we filter out the missing emotions from ArmanEmo so that the models can predict only joy, anger, fear, and sadness. Moreover, to compare the performance of XLM-EMO to other models in our baselines, we also train and evaluate another XLM-EMO model considering all seven emotion classes in our original dataset.

To fine-tune the pre-trained language or emotion models for our task, i.e., Emotion Detection, we add a fully-connected dense layer on top of these models. We also introduce a cross-entropy loss function to perform the multi-label classification task using the resulting models. While fine-tuning, we freeze all the networks' weights except the weights for the last layer, i.e., the added dense layer.

\subsubsection{Data Pre-processing}

After splitting the data into training and testing sets, we follow some pre-processing steps to transform the data into a format proper to feed into the final model. As the first pre-processing step, we normalize the text using Parsivar \cite{mohtaj2018parsivar}, a toolkit for Persian text pre-processing. It applies some rule-based space correction steps (including word, punctuation, and affix spacing), along with some character refinement operations (such as removing stretching letters). Its normalization rules, however, are not comprehensive. For example, "Arabic Sukun" will not be removed after normalization. Besides, we need to consider some task and domain-specific rules while pre-processing the text from social media. So, after introducing the text to the Parsivar normalizer, we perform some additional pre-processing steps, which include removing:

\begin{enumerate}
  \item any English character from the text
  \item letters repeated more than twice in the non-standard spelling of Persian words which are often intentionally used for more emphasis in the informal text (like \RL{خيييلللييی} instead of \RL{خيلی})
  \item any Arabic diacritics from the text which are not removed by Parsivar normalizer
  \item any remaining non-Persian characters after performing the above steps
  \item the hashtag sign (“\#”) from the text while keeping the information included in the hashtags
  \item Persian numeric characters from the text

\end{enumerate}


\subsubsection{Hyperparameters}

When fine-tuning the pre-trained ParsBERT model, we used the same hyperparameters set by Farahani et al. except for the batch size, maximum sequence length, and the number of epochs. We fine-tuned the ParsBERT on our training dataset for eight epochs using a batch size of 32 while limiting the maximum sequence length to 128. We used AdamW (Adam with decoupled weight decay) optimizer with \(\beta_1=0.9\), \(\beta_2=0.999\), \(\lambda=0.01\). We also utilize a learning rate scheduler, linearly decreasing from the initial learning rate (2e-5) to 0 by the end of the last epoch. For other Language Models, we have utilized the default hyperparameters used in the open-source library Transformers \cite{wolf-etal-2020-transformers}.

\subsection{Results}

In this section, we discuss the results of different experiments performed to showcase the quality of ArmanEmo. The first part of this section summarizes the performance of various deep models on ArmanEmo using different evaluation metrics. The results for the Zero-shot tests using XLM-Emo on the dataset are discussed next. Lastly, we compare the generalization of ArmanEmo against EmoPars in a transfer learning setting.

\subsubsection{Comparison of baseline models on ArmanEmo}

In order to compare the emotion models based on fine-tuned Language Models in our baselines against other deep neural networks, we trained and evaluated CNN-based and RNN-based models on our train and test datasets. It is worth noting that for all the models that are not based on language models, we used the same pre-trained word vectors as the embeddings, which are trained on Common Crawl and Wikipedia using fastText \cite{grave2018learning}. As Table \ref{tab:modelscomparison} shows, the fine-tuned XML-RoBERTa-large model significantly outperforms other models on our test set in terms of average macro F1 score, precision, and recall. Moreover, it can be seen that there is a performance gap between fine-tuned LMs and RNN/CNN-based models in emotion detection. This significant performance difference can be related to the superior ability of pre-trained Language Models in extracting rich and non-trivial knowledge from the textual data.

\begin{table}[H]
 \caption{Comparison between the performance of different DNN models and Language Models}
  \centering
  \begin{tabular}{llll}
    \toprule
    Model    & Precision (Macro)    & Recall (Macro)   & F1  (Macro) \\
    \midrule
    FastText \cite{joulin2016bag} & 54.82  & 46.37 & 47.24   \\
    HAN \cite{yang-etal-2016-hierarchical} & 49.56 & 44.12 & 45.10      \\
    RCNN \cite{lai2015recurrent} & 50.53 & 48.11 & 47.95      \\
    RCNNVariant & 51.96 & 48.96 & 49.17  \\
    TextAttBiRNN \cite{bahdanau2014neural, raffel2015feed} & 54.66 & 46.26 & 47.09 \\
    TextBiRNN & 51.45 & 47.16 & 47.14 \\
    TextCNN \cite{kim-2014-convolutional} & 58.66 & 51.09 & 51.47  \\
    TextRNN \cite{liu2016recurrent} & 49.39 & 47.20 & 46.79  \\
    ParsBERT & 67.10 & 65.56 & 65.74 \\
    XLM-Roberta-base & 72.26 & 68.43 & 69.21 \\
    XLM-Roberta-large & \textbf{75.91} & \textbf{75.84} & \textbf{75.39} \\
    XLM-EMO-t & 70.05 & 68.08 & 68.57 \\
    
    \bottomrule
  \end{tabular}
  \label{tab:modelscomparison}
\end{table}

\begin{figure}[H]
\centering
\includegraphics[width=8cm]{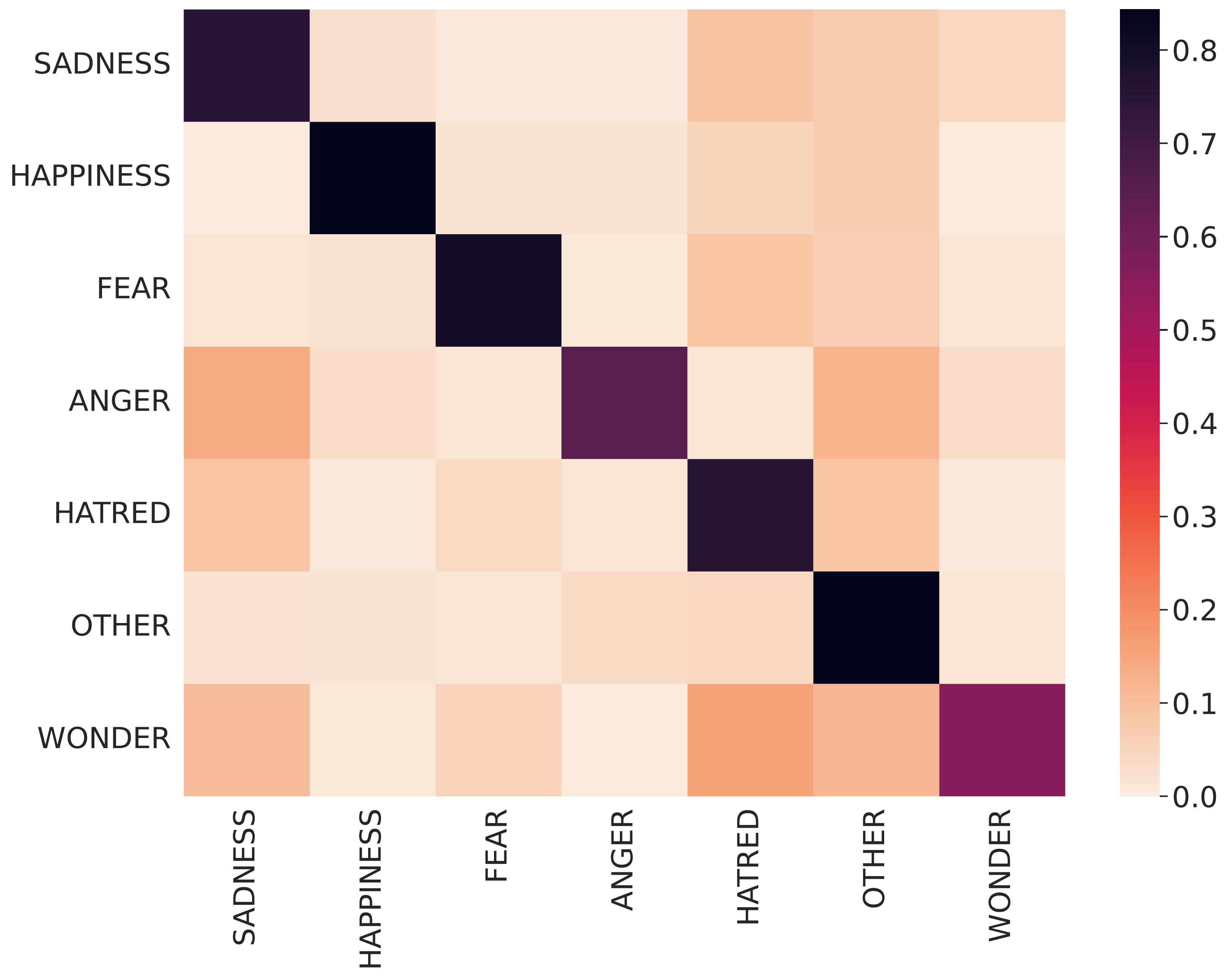}
\caption{Confusion Matrix for XLM-RoBERTa-large Predictions}
\label{fig:confusion}
\end{figure}

The performance of the best model among our baseline models (XLM-RoBERTa-large) is summarized in Table \ref{tab:xlmrobertalargeevaluation} and Figure \ref{fig:confusion}. The model achieves a macro average F1 score of 75.39 on the test set. It presents the best performance on emotions like Happiness, Fear, and Sadness. On the other hand, it obtains the lowest F1 score on emotions like Anger, Other, and Hatred.

\begin{table}[H]
 \caption{Evaluation Metrics Resulted from The Best Model (XLM-RoBERTa-large)}
  \centering
  \begin{tabular}{llllll}
    \toprule
    Emotion & Precision     & Recall    & F1 & Support (No. of Test Examples)  \\
    \midrule
    Anger & 74.62 & 62.99 & 68.31 & 154   \\
    Fear & 78.69 & 84.21 & 81.36 & 57  \\
    Happiness & 86.02 & 87.27 & 86.64 & 275  \\
    Hatred & 63.64 & 75.38 & 69.01 & 65 \\
    Other & 60.98 & 77.72 & 68.34 & 193  \\
    Sadness & 82.45 & 77.10 & 79.68 & 262  \\
    Wonder & 84.96 & 66.21 & 74.42 & 145 \\
    \midrule
    Macro Average & 75.91 & 75.84 & 75.39 & 1151 \\
    \bottomrule
  \end{tabular}
  \label{tab:xlmrobertalargeevaluation}
\end{table}

We manually investigated the mislabeled sentences to better analyze the situations where the best model is performing poorly. Table \ref{tab:erroranalysis} presents some randomly selected samples of these mislabeled sentences generated by the best model. Going through some of these examples, what can be inferred is that the model outputs the wrong label whenever a given sentence carries mixed emotions. In such situations, the assignment of one and only one exact emotion to the sentence might be challenging even for humans. That is why multi-label classifiers are used against multi-class classifiers to include the presence of more than one emotion in a given sentence. Since we are dealing with a multi-class classification problem in this study, we presume that each sentence carries only one emotion; hence, it is assumed that there is only one true target for each sentence. However, this might not be the case for all situations. One can judge, based on the given sentences, that some of the model's predictions are not irrelevant at all. In fact, depending on the context, these predicted labels might be considered as valid as the assigned ground truths. However, there are still other situations where the model's poor performance seems to be related to the model being biased to the occurrence of some specific words or combinations of words in the sentence.

\begin{table}[H]
 \caption{Randomly-selected samples of mislabeled sentences (generated by XLm-RoBERTa-large)}
  \centering
  \begin{tabular}{lll}
    \toprule
    Sentence & Ground Truth Label & Model Prediction \\
    \midrule

    \scriptsize{\RL{واقعا حال به هم زنه اين حجم از داستان سرايی}} & Hatred & Wonder \\
    \scriptsize{\RL{درباره تجاوز يا چيزهاي شبيه به اون برا جذب لايك و توجه}} & & \\
    & & \\
    The amount of fake stories being spread about & & \\
    rape or things like that for likes & & \\
    and attention is honestly revolting. & & \\
    \midrule
    \scriptsize{\RL{کتاب امروز به دستم رسيد. جنس برگه هاش خوب بودن. رنگهای شادی داشت.}} & Happiness & Other \\
    \scriptsize{\RL{اسم شخصيتهای داستانها خنده دار و جالب  بود. داستانهاش هم تکراری نبودن.}} & & \\
    \scriptsize{\RL{فقط چون کل سی جلد يکجا جمع شده کمی دست رو خسته ميکنه}} & & \\
    & & \\
    I got my hands on the book today. The paper has good quality. & & \\
    A lot of vibrant colors. The characters in each story had & & \\
    funny and interesting names. The stories weren’t repetitive. & & \\
    Although because it was a collection of & & \\
    all 30 books your hands would get tired. & & \\
    \midrule
    \scriptsize{\RL{انصافا چقدر فک کردی اينو نوشتی خخخ! ولی احسنت جالب بود}} & Happiness & Wonder \\
    & & \\
    Honestly how much thought  did you put in when & & \\
    you wrote this, LOL! But nice, it was interesting & & \\
    \midrule
    \scriptsize{\RL{بريم يک سيگاری بکشيم شايد حال داد.}} & Happiness & Other \\
    & & \\
    Lets go smoke a cigarette, it might be fun. & & \\
    \midrule
    \scriptsize{\RL{چجوری ميشه آدم خوشحالو خندون باشه بعد يه دفعه غمگين بشه؟}} & Sadness & Wonder \\
    \scriptsize{\RL{فکنم دچار نوع خاصی از خود درگيری شدم}} & & \\
    & & \\
    How can a person be happy and suddenly feel so down? & & \\
    I think I am struggling with a specific kind of internal conflict. & & \\
    \midrule
    
    \scriptsize{\RL{ بزرگترين گناه ترس هست و اين خانم شجاعترين مرد ميدان بود  درود بر شرفش}} & Happiness & Fear \\
    & & \\
    The biggest sin is fear and this woman was the bravest hero. & & \\
    Peace be upon her. & & \\
    \midrule
    \scriptsize{\RL{سلبريتيا ميان از وضعيت بد اقتصادی مردم انتقاد ميکنن بعد }} & Wonder & Anger \\
    \scriptsize{\RL{بليط تئاتر خودشون صد و هشتاد هزار تومنه :))}} & & \\
    & & \\
    The celebrities want to complain about the bad economic times & & \\
    yet they make their own theater shows cost 180 Tomans :)) & & \\
    \midrule
    \scriptsize{\RL{ موسيقی و شاد بودن حرام نيست}} & Other & Happiness \\
    & & \\
    Listening to music and being happy isn’t haram. & & \\
    \midrule
    \scriptsize{\RL{در مورد راه شيری چیز خاصی نميشه گفت حيرت انگيزه!}} & Wonder & Happiness \\
    \scriptsize{\RL{ولی از خفن بودن تلسکوپ هابل هم نميشه گذشت}} & & \\
    & & \\
    I cant find the words to describe the Milky Way & & \\
    since it is truly breathtaking! But that also doesn’t take away & & \\
    from how amazing the Hubble telescope is. & & \\
    \midrule
    \scriptsize{\RL{ملت رفتن زير آخرين پستهای اينستاگرام مرحوم گفتن روحت شاد.}} & Wonder & Sadness \\
    & & \\
    Everyone's gone under the decedent’s most recent post & & \\
    on Instagram and have said rest in peace. & & \\

    
    \bottomrule
  \end{tabular}
  \label{tab:erroranalysis}
\end{table}

\subsubsection{Zero-shot tests using XLM-EMO}

Table \ref{tab:zeroshottest} compares the performance of XLM-EMO (a multilingual emotion model) as a zero-shot classifier against another version of it which is trained on our dataset (while the data is limited to only four basic emotions, as is explained in section \ref{subsec:32}) (the F1 score for the XLM-RoBERTa-large model and ParsBERT are also provided for comparison). As expected, the model trained on our train set has a superior performance in terms of macro-averaged F1 score. However, the performance of XLM-EMO as a zero-shot classifier, which has not seen any of our Persian text during its training, is still impressive. It's performance is somewhat comparable to the performance of ParsBERT, which has been pre-trained on large Persian corpus. This result suggests that using a multilingual emotion model can be very helpful in detecting emotion from Persian text in the lack of any emotion dataset.

\begin{table}[H]
 \caption{Result for zero-shot test using XLM-EMO while considering four basic emotions}
  \centering
  \begin{tabular}{ll}
    \toprule
    Model & F1 Score \\
    \midrule
    XLM-EMO Zero-Shot & 75.28 \\
    XLM-EMO Trained & 84.46 \\
    XLM-RoBERTa-large & 85.02 \\
    ParsBERT & 79.38 \\

    \bottomrule
  \end{tabular}
  \label{tab:zeroshottest}
\end{table}

\subsubsection{Comparing emotion datasets for Persian text}
\label{subsubsec:333}

To better demonstrate the capabilities of our emotion dataset, we compare the performance of the best model in our baselines (XLM-RoBERTa-large) when it is trained on ArmanEmo against when it is trained on EmoPars. Before this experiment, some preparatory steps need to be done on EmoPars since the sentences in this collection are released with no major emotion specified by the authors. Instead, Sabri et al. have directly published the results of their crowd-sourced data labeling procedure. Each sample in EmoPars contains a sentence along with six numbers (vote counts in the range of [0,5]) corresponding to each of the six emotions considered in this study. Selecting the final emotion for each sentence can be done by introducing a threshold, a task delegated to the users of EmoPars. For our purposes, we decide to remove samples with no dominant emotion, and, to do this, we define a dominance threshold which is set to 3. In other words, we filter out the sentences for which none of the six emotions has received a YES vote from 3 out of 5 annotators. For the remaining samples, if the sentence has more than one dominant emotion, it gets removed again. Otherwise, the dominant emotion will be selected as the final label of the given sentence. Following this procedure, we selected 5477 tweets (out of 30000 samples) from EmoPars. Since the original dataset is not split into train and test sets, we randomly selected 80 percent of these tweets as the train set and the remaining 20 percent as the test set.

To make a fair comparison, we run our experiments in four different situations. In the first two combinations, we use XLM-RoBERTa-large, which is trained on our train set, and in the last ones, we use XLM-RoBERTa-large, which is trained on the train set of EmoPars. For both cases, we evaluate the model's performance on our test set and on the test set of EmoPars. The results of this study (in terms of macro-averaged F1 score) are summarized in Table \ref{tab:comparisonwithemopars}

\begin{table}[H]
 \caption{Macro-averaged F1 score results for comparison between ArmanEmo and EmoPars}
  \centering
  \begin{tabular}{lll}
    \toprule
    Train Set    & Tested on EmoPars   & Tested on ArmanEmo \\
    \midrule
    ArmanEmo & \textbf{26.68} & \textbf{75.39}  \\
    EmoPars & 6.96 & 7.16  \\
    
    \bottomrule
  \end{tabular}
  \label{tab:comparisonwithemopars}
\end{table}

As it can be seen in Table \ref{tab:comparisonwithemopars}, the model that has been trained on our train set has a superior performance to the model trained on the train set of EmoPars, even when the models are evaluated on the test set of EmoPars. When testing both models on the test set of EmoPars, the F1 score is more than 19 percent higher for the model trained on ArmanEmo. This observation indicates that the generalizability of ArmanEmo is superior to that of EmoPars.

The noticeably poor performance of the models that were trained on EmoPars can be related to the fact that this dataset is labeled through crowd-sourcing, consisting of many noisy labels. In our study, on the other hand, the labels were selected according to a meticulously designed procedure discussed in section \ref{subsec:31}. We have chosen the labels with much fewer errors, so it was expected that ArmanEmo would have less noise and, therefore, better generalizability.

\section{Conclusion}
\label{sec:4}

In recent years, emotion detection from text has been a topic of interest to researchers in natural language processing. Among different approaches to developing emotion detection systems, data-driven methods in general, and deep learning algorithms in particular, have received considerable attention in this field. A significant issue with deep learning algorithms in emotion detection from text is the lack of annotated labels. In this study, we provide a manually annotated dataset suitable for training deep learning algorithms. To demonstrate the high quality of our dataset, we build several strong baseline models, including state-of-the-art language models. Through transfer learning experiments, we show the superior generalizability of our proposed dataset. The final dataset, containing more than 7000 samples, is publicly available for non-commercial use.

\section*{Acknowledgement}

The authors would like to thank \href{https://aipaa.ir/}{Arman Rayan Sharif} for providing required financial and computational resources in this project.

\bibliographystyle{unsrt}  
\bibliography{references}

\end{document}